\documentclass[11pt]{article}

\usepackage[final]{acl}

\usepackage{latexsym}
\usepackage{microtype}
\usepackage{fontspec}
\newfontfamily\jpfont{HaranoAjiMincho}[Extension=.otf, UprightFont=*-Regular, BoldFont=*-Bold] % Japanese; ships with TeX Live
\usepackage{inconsolata}
\usepackage{graphicx}
\usepackage{booktabs}
\usepackage{float}
\usepackage{multirow}
\usepackage{array}
\usepackage{amsmath}
\usepackage{amssymb}
\usepackage{bm}
\title{PAINT: A Framework for Multilingual Pain Representation from Medical Licensing Exams to Real-World Clinical Texts}
\author{%
  \begin{minipage}{\textwidth}\centering
  \mbox{Boya~Zhang$^{1,2,*}$}, \mbox{Yuanchao~Ma$^{3,*}$}, \mbox{Jiageng~Wu$^{4,5,*}$},\\
  \mbox{Krittaphas~Chaisutyakorn$^{2,6,7}$}, \mbox{Rodrigo~Gameiro$^{2,4}$}, \mbox{Ahram~Han$^{2,8}$},\\
  \mbox{Chien-Yu~Huang$^{3,9}$}, \mbox{Nikhil~Jaiswal$^{2,3}$}, \mbox{Marianna~Leite$^{10,11}$},\\
  \mbox{Miriam~Mora-Mau$^{12}$}, \mbox{Jaime~Pineda-Moreno$^{13}$}, \mbox{Siena~Placino$^{14}$}, \mbox{Ryohei~Yamamoto$^{2,15}$},\\
  \mbox{Yugang~Jia$^{2}$}, \mbox{Jie~Yang$^{4,5}$}, \mbox{Leo~Anthony~Celi$^{2,4,6,16}$}, \mbox{Douglas~Teodoro$^{1}$}\\[4pt]
  {\normalfont
  \mbox{$^{1}$University of Geneva}, \mbox{$^{2}$Massachusetts Institute of Technology}, \mbox{$^{3}$McGill University},
  \mbox{$^{4}$Harvard Medical School}, \mbox{$^{5}$Brigham and Women's Hospital},
  \mbox{$^{6}$Harvard T.H.\ Chan School of Public Health}, \mbox{$^{7}$Mahidol University},
  \mbox{$^{8}$Seoul National University Hospital}, \mbox{$^{9}$National Taiwan University},
  \mbox{$^{10}$Universidade de S\~ao Paulo}, \mbox{$^{11}$Santa Marcelina College of Medicine},
  \mbox{$^{12}$Universidad Peruana Cayetano Heredia}, \mbox{$^{13}$HM Hospitales},
  \mbox{$^{14}$St.\ Luke's Medical Center College of Medicine}, \mbox{$^{15}$Fukushima Medical University},
  \mbox{$^{16}$Beth Israel Deaconess Medical Center}\\[2pt]
  $^{*}$Equal contribution\\[2pt]
  \url{https://github.com/tinaboya/PAINT}}
  \end{minipage}}

\begin{document}
\maketitle

\begin{abstract}
Pain is one of the most common and clinically important symptoms, reflecting disease onset, progression, and treatment response. 
However, pain is difficult to accurately capture and interpret from medical text, as its expression varies across languages, documentation types, and patient demographics, and is often embedded in terminology, negation, temporality, and contextual descriptions. Existing pain-related studies remain largely monolingual or limited to narrow clinical settings.
In this study, we introduce PAINT, a multilingual framework for studying pain representation across medical exams and real-world clinical texts. 
PAINT includes nine pain-focused dataset partitions spanning English, Spanish, Portuguese, and Japanese, with clinician annotations in all four languages.
We curated a structured schema covering pain relevance, pain presence, anatomical location, pain descriptors, and demographic information. 
To capture multilingual pain representations, we leverage large language models (LLMs) and adopt an ensemble strategy to improve performance.
Ten clinicians across four languages validate the labels: the ensemble reaches macro-F1 of 0.91--1.00 on pain relevance and pain presence and 92--100\% accuracy on demographics, with inter-annotator $\kappa$ of 0.82--1.00. We release the WorldMedQA-Pain ensemble labels, the extraction pipeline, validation scripts, and clinician annotation sheets for publicly licensed corpora, providing a foundation for pain-centered clinical NLP and for studying pain documentation across languages and healthcare contexts.

% The representation of pain changes across individuals, languages, and the healthcare systems that treat us. Medical licensing exams and clinical notes capture pain from what medicine teaches in theory, and what clinicians encounter in practice. We introduce \textbf{PAINT}, the first multilingual framework for cross-corpus comparison of pain representation between medical education and clinical practice across languages. We build the exam set \textbf{WorldMedQA-Pain} and the note set \textbf{BRIDGE-Pain} in Portuguese, Japanese, Spanish, and English using an ensemble of three open-weight language models, and validate them on physician annotations. We analyze pain representation across pain presence rate, pain location, and demographic information. We find that pain is a more frequent chief complaint in real-world clinical notes than in medical licensing exams, general health queries, and medication safety incident reports. We demonstrated that the language models in the \textbf{PAINT} framework are valid for creating a pain representation corpus and can be extended to additional languages and sets. 

\end{abstract}

\begin{figure}[!t]
\centering
\includegraphics[width=\columnwidth]{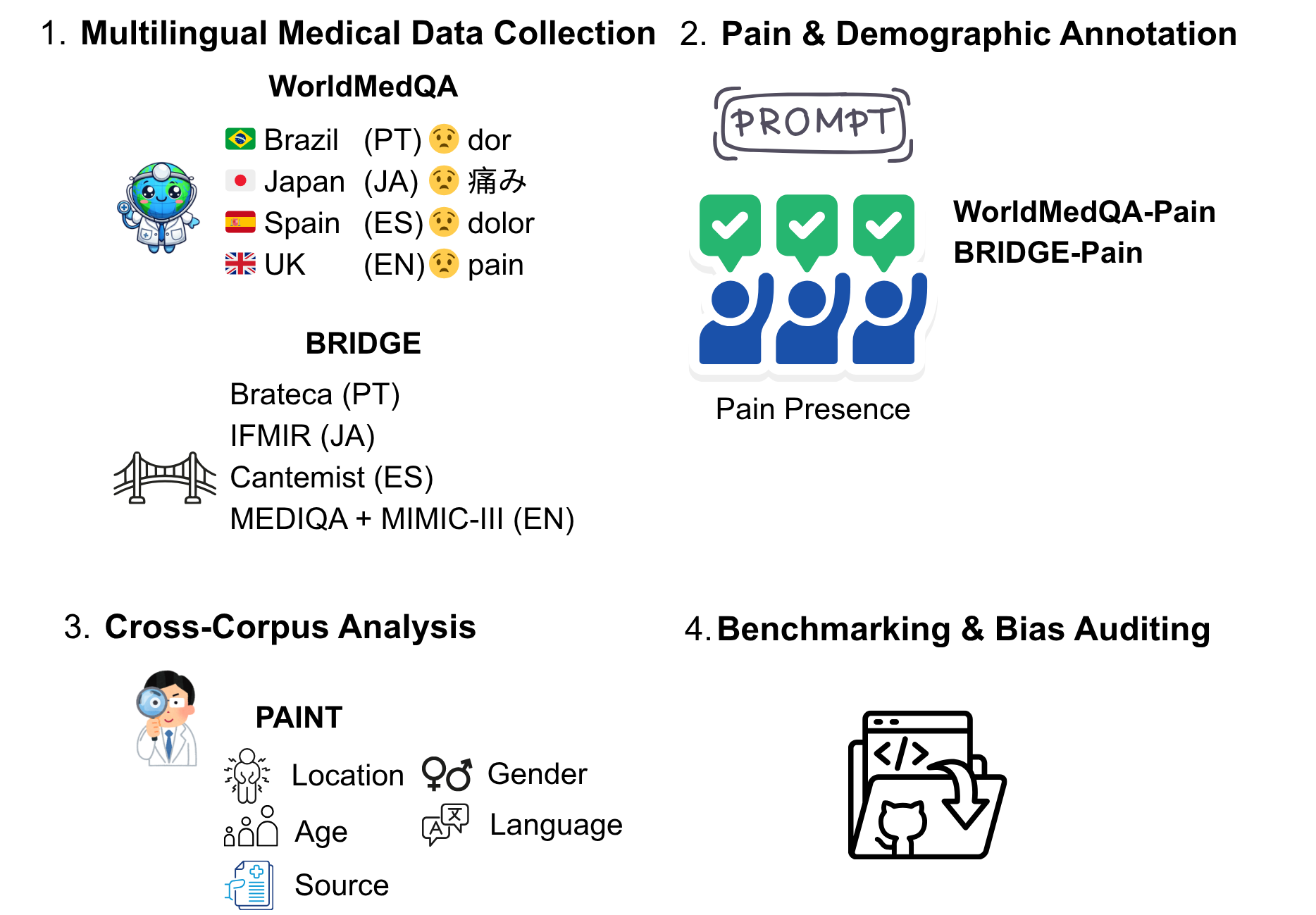}
\caption{Overview of PAINT: medical licensing examinations and clinical text in four languages, annotated with an ensemble of open-weight language models and validated by clinicians.}
\label{fig:pipeline}
\end{figure}

\section{Introduction}
\label{sec:intro}

Pain is a common and consequential symptom in clinical care, and prior work has shown that its assessment, treatment, and documentation vary with patient race, gender, age, and language \citep{singhal2016racial, hoffmann2001girl, eccleston2021pediatric, rambachan2023paindisp, lor2022pain}. Such variation may affect clinical decision making. Studying these differences across languages, document types, and patient demographics is therefore relevant both to clinical practice and to the evaluation of robustness and fairness in large language model (LLM)-based clinical NLP systems.

Multilingual clinical NLP primarily relies on two types of resources. Medical licensing examination benchmarks, such as MedQA and WorldMedQA, evaluate structured clinical knowledge in standardized exam settings \citep{jin2021medqa, alonso2024medexpqa, worldmedqa2024}. Real-world clinical corpora, including MIMIC and BRIDGE, capture how clinicians document patient encounters in routine practice \citep{johnson2016mimic, bridge2025}. The two offer different views of medical language: one reflects what medicine teaches, the other how medicine is practiced.

Despite growing interest in pain-related NLP, available resources remain limited in both linguistic and clinical diversity. A recent systematic review identified small sample sizes, monolingual datasets, and the underrepresentation of non-English populations as persistent challenges in the field \citep{rajwal2024decade}. Existing annotated pain corpora have similarly focused on English clinical text, including UK mental health records annotated for pain relevance, anatomical location, and pain character \citep{chaturvedi2023pain}, and US clinical notes annotated for chronic pain episodes, causes, locations, severities, and medications \citep{carlson2020chronic}. To date, no resource provides a unified multilingual representation of pain across both medical education and clinical practice.

In this work, we present PAINT, to our knowledge the first multilingual framework for analyzing pain representation across Portuguese, Japanese, Spanish, and English. PAINT is a data resource, and the LLM ensemble serves as its labeling methodology, with open-weight models benchmarked against clinician-annotated labels. We began from a clinical question, how pain is documented and how that documentation changes with language and patient demographics, and selected corpora accordingly. Cross-language corpora with matched documentation purpose for pain are scarce, so we work with the real-world clinical text that is available, in which pain appears incidentally rather than by design. PAINT comprises two resources: WorldMedQA-Pain, containing 2,963 medical licensing examination questions, and BRIDGE-Pain, containing 17,795 real-world clinical records. Pairing exam and clinical text within each language under a common schema, with clinician validation across all four languages, enables within-language comparisons between exam and clinical text. Figure~\ref{fig:pipeline} provides an overview of the PAINT framework.

Our contributions are threefold:

(1) \textbf{Resource.} A multilingual pain dataset that pairs medical-licensing-exam text with real clinical notes under one structured schema across four languages, with clinician-validated labels and a released extraction pipeline. The validated labels also serve as an evaluation set for pain extraction, useful for testing clinical NLP systems and LLMs, especially in non-English settings where such data are scarce.

(2) \textbf{Methodology.} A reproducible, clinician-validated labeling pipeline with three methodological strengths: generalizability, a cross-lingual, cross-corpus pain extraction and labeling framework; transferability, the pipeline validated on WorldMedQA-Pain transfers to BRIDGE-Pain, where clinician review confirms macro-F1 of at least 0.91 for pain relevance and pain presence and at least 92\% accuracy for demographics across all five clinical corpora; and efficiency, the open-weight ensemble matches GPT-5 on the consensus subset and runs locally on a single GPU, making it usable on controlled-access clinical corpora and in resource-constrained settings.

(3) \textbf{Documentation.} We document the released labels per partition: sample sizes, consensus coverage, pain relevance and presence rates, and demographic field coverage, together with the annotation protocol, clinician instructions, and every inter-annotator disagreement, so that users can judge the reliability of each field and partition before reusing them.

\section{Related Work}
\label{sec:related}

\paragraph{Pain disparities and clinical bias.}
Recent studies suggest that AI systems may reproduce the disparities in pain assessment and treatment documented across race, gender, age, and language \citep{singhal2016racial, hoffman2016racial, hoffmann2001girl, eccleston2021pediatric, rambachan2023paindisp}. In knee radiography, radiologist-based severity grading explains only a small share of racial disparities in reported pain, whereas a model trained on patients' own pain reports explains substantially more of them \citep{pierson2021algorithmic}; demographic bias has also been observed in LLM-assisted opioid prescribing and pain management \citep{omar2026sociodemographic, omar2025review}. Beyond clinical prediction, \citet{berke2026pain} propose a computational theory of intuitive pain reasoning, modeling how humans infer and describe pain through mechanisms analogous to intuitive physics and theory of mind.

\paragraph{Multilingual clinical resources.}
Examination datasets such as MedQA \citep{jin2021medqa}, MedMCQA \citep{pal2022medmcqa}, MedExpQA \citep{alonso2024medexpqa}, and WorldMedQA-V \citep{worldmedqa2024} evaluate clinical knowledge across one or multiple languages. In parallel, clinical corpora such as MIMIC-III and MIMIC-IV \citep{johnson2016mimic, johnson2023mimic4} and the ICU databases unified by ricu \citep{pollard2018eicu, faltys2021hirid, thoral2021amsterdamumc, bennett2023ricu} provide large-scale records of routine clinical documentation. More recently, BRIDGE \citep{bridge2025} has expanded clinical benchmarking across institutions and languages, and DRAGON \citep{bosma2025dragon} across Dutch clinical NLP tasks. However, none of these resources represents pain as a standardized, structured variable suitable for cross-lingual analysis.

\paragraph{LLM-based clinical information extraction.}
Few-shot prompting makes large language models effective clinical information extractors without task-specific fine-tuning \citep{agrawal2022large}, and chain-of-thought prompting \citep{wei2022cot} and self-consistency decoding \citep{wang2023selfconsistency} further improve their reasoning. Ensemble reasoning enhances the stability and consistency of structured clinical extraction \citep{chang2024ensemble}. PAINT builds on this work by using two multilingual chain-of-thought prompting pipelines and requiring agreement among three models to generate high-precision silver annotations at scale.

\section{Multilingual Medical Data Collection}
\label{sec:method}
\label{sec:data}

\paragraph{WorldMedQA-Pain.}\label{sec:data:worldmedqa}
WorldMedQA-Pain is constructed from the text-only national licensing examination partitions assembled by the WorldMedQA team as an extension of WorldMedQA-V \citep{worldmedqa2024}, retaining four partitions in their original languages. The resulting corpus contains 2,963 instances spanning Brazilian Portuguese, Japanese, Spanish, and English. The four partitions correspond to REVALIDA (Brazil), the National Medical Licensing Examination (Japan), MIR (Spain), and PLAB (United Kingdom). Each instance carries clinician-annotated labels for pain relevance and pain presence, produced under the multilingual annotation protocol described in Appendix~\ref{app:annotation}.

\paragraph{BRIDGE-Pain.}
BRIDGE-Pain is derived from the multilingual BRIDGE benchmark \citep{bridge2025} and consists of five clinical-text partitions: MEDIQA 2023 Chat-A (English doctor--patient dialogues) \citep{dataset-MTS-Dialog-MEDIQA-2023}, MIMIC-III Outcome (English ICU discharge summaries) \citep{dataset-MIMIC-III-Outcome}, Cantemist (Spanish oncology reports) \citep{dataset-cantemist-2020}, Brateca (Brazilian Portuguese electronic health records) \citep{dataset-Brateca}, and IFMIR (Japanese medication-safety incident reports) \citep{dataset-ifmir-wong-2024}. We retain the complete MEDIQA and Cantemist partitions and randomly down-sample MIMIC-III, Brateca, and IFMIR to 5,000 records each. The final BRIDGE-Pain corpus contains 17,795 clinical documents, all annotated with the same multilingual ensemble pipeline and validated by clinicians on sampled records.

\section{Pain and Demographic Annotation}
\label{sec:annotation}
\label{sec:method:pipeline}

We extract pain and demographic information using two chain-of-thought prompting pipelines \citep{wei2022cot, kojima2022zeroshot}, one for pain attributes and one for demographic attributes. All source texts are processed in their original languages. Hierarchical constraints on the pain labels are enforced in the prompt. The complete prompts are provided in Appendix~\ref{app:prompts}.

\paragraph{Annotation schema.}
Each record is described by five fields organized under a hierarchical schema. \emph{Pain relevance} indicates that the text mentions pain in any way, including a pain term, synonym, or derivative (e.g., \emph{painful}, \emph{painless}, or \emph{painkiller}), as well as an explicit denial of pain. \emph{Pain presence}, conditioned on relevance, indicates that pain is asserted to be present in the patient at least once. A record in which the only pain mention is a denial or refers to an attribute of an intervention is therefore relevant but not pain-present. \emph{Anatomical location}, conditioned on pain presence, records the anatomical site or sites of pain as stated in the text, in concise medical terminology, or \emph{unspecified} when pain is present but no site is named. \emph{Pain descriptors} capture free-text descriptions of pain quality. \emph{Demographics} include age, gender, ethnicity, and occupation when stated in the text. The distinction between relevance and presence is the central design choice of the schema: it separates records that mention pain from those in which pain is actually present, ensuring that denials and intervention-related mentions are not counted as pain. Table~\ref{tab:mwp-examples} gives examples of such mentions in each language.

\begin{table}[t]
\centering
\footnotesize
\setlength{\tabcolsep}{4pt}
\begin{tabular}{@{}lp{0.40\columnwidth}l@{}}
\toprule
\textbf{Lang} & \textbf{Pain mention in source} & \textbf{Reason} \\
\midrule
en & \emph{``She has no pain''} & denial \\
es & \emph{``sin fiebre ni dolor''} (no fever or pain) & denial \\
pt & \emph{``n\'odulo indolor''} (painless nodule) & \emph{painless} \\
ja & {\jpfont 右陰囊の無痛性腫大} (painless swelling) & {\jpfont 無痛性} (\emph{painless}) \\
\bottomrule
\end{tabular}
\caption{Examples where pain is mentioned but not asserted in the
patient, the mention-without-presence case, agreed by all three
models. The trace is an explicit denial or a \emph{painless}-type
derivative. Quotes are excerpts from the source text.}
\label{tab:mwp-examples}
\end{table}

\paragraph{Model selection.}
We evaluate five open-weight LLMs on WorldMedQA-Pain using native-language inputs, with GPT-5 (\texttt{gpt-5-2025-08-07}) included as a proprietary reference. As shown in Table~\ref{tab:bench}, Gemma-4-E4B (4B), Qwen3-8B (8B), and DeepSeek-R1-Distill-Qwen-32B (32B) perform best overall, with macro-F1 scores ranging from 0.87 to 0.97 for pain relevance and pain presence across the four partitions. LLaMA-3.1-8B and MedGemma-4B score lower across the partitions and are not included in the ensemble.

\begin{table*}[t]
\centering
\small
\setlength{\tabcolsep}{4pt}
\begin{tabular}{lcccccccc}
\toprule
& \multicolumn{2}{c}{\textbf{Brazil}}
& \multicolumn{2}{c}{\textbf{Japan}}
& \multicolumn{2}{c}{\textbf{Spain}}
& \multicolumn{2}{c}{\textbf{UK}} \\
\cmidrule(lr){2-3}\cmidrule(lr){4-5}\cmidrule(lr){6-7}\cmidrule(lr){8-9}
\textbf{Model}
  & \textit{Rel.} & \textit{Pres.}
  & \textit{Rel.} & \textit{Pres.}
  & \textit{Rel.} & \textit{Pres.}
  & \textit{Rel.} & \textit{Pres.} \\
\midrule
& \multicolumn{8}{c}{\textit{Independent evaluation}} \\
\cmidrule(lr){2-9}
Gemma-4-E4B
  & 0.894 & 0.872 & 0.972 & 0.951 & 0.905 & 0.921 & 0.946 & 0.936 \\
Qwen3-8B
  & 0.879 & 0.873 & 0.948 & 0.956 & 0.913 & 0.917 & 0.931 & 0.932 \\
DeepSeek-R1-Distill-Qwen-32B
  & 0.870 & 0.872 & 0.936 & 0.965 & 0.889 & 0.892 & 0.935 & 0.949 \\
\cmidrule(lr){1-9}
LLaMA-3.1-8B
  & 0.668 & 0.744 & 0.657 & 0.751 & 0.674 & 0.738 & 0.718 & 0.810 \\
MedGemma-4B
  & 0.685 & 0.730 & 0.786 & 0.782 & 0.791 & 0.808 & 0.810 & 0.824 \\
\cmidrule(lr){1-9}
GPT-5$^\dag$
  & \textbf{0.910} & \textbf{0.892} & \textbf{0.993} & \textbf{0.970} & \textbf{0.927} & \textbf{0.929} & \textbf{0.954} & \textbf{0.951} \\
\midrule
& \multicolumn{8}{c}{\textit{Ensemble on the consensus subset}} \\
\cmidrule(lr){2-9}
\textit{Coverage}
  & \multicolumn{2}{c}{88.9\%}
  & \multicolumn{2}{c}{92.9\%}
  & \multicolumn{2}{c}{92.4\%}
  & \multicolumn{2}{c}{91.0\%} \\
\cmidrule(lr){2-3}\cmidrule(lr){4-5}\cmidrule(lr){6-7}\cmidrule(lr){8-9}
3-model ensemble
  & 0.922 & \textbf{0.916}
  & 0.992 & \textbf{0.992}
  & 0.935 & \textbf{0.960}
  & 0.961 & 0.969 \\
GPT-5$^\dag$
  & \textbf{0.925} & 0.915 & \textbf{0.995} & 0.989 & \textbf{0.937} & 0.955 & \textbf{0.966} & \textbf{0.978} \\
\bottomrule
\end{tabular}
\caption{Macro F1 for pain relevance and pain presence on
WorldMedQA-Pain. ``Rel.''\ and ``Pres.''\ denote pain relevance and
pain presence, respectively. \textit{Independent} reports each model
evaluated on the full corpus. \textit{Ensemble} reports the ensemble,
and GPT-5 for comparison, evaluated on the consensus subset,
consisting of records on which all three selected models agree on both
pain fields. \textit{Coverage} is the fraction of the full corpus
retained in the consensus subset. GPT-5$^\dag$ is included as a
proprietary reference.}
\label{tab:bench}
\end{table*}

\paragraph{Ensemble strategy.}\label{sec:method:ensemble}\label{sec:method:models}
PAINT combines the three highest-performing open-weight models into a consensus ensemble; performance on non-consensus rows is reported in Table~\ref{tab:abstain}.

Let $y_i^{(m)}$ denote the prediction of model $m\in\{1,2,3\}$ for instance $i$, and let $\bot$ denote abstention. For the binary tasks of pain relevance and pain presence, we use a strict three-way agreement rule:
\begin{equation}
\hat{y}_i =
\begin{cases}
y_i^{(1)}, & \text{if } y_i^{(1)} = y_i^{(2)} = y_i^{(3)},\\
\bot, & \text{otherwise}.
\end{cases}
\label{eq:strict}
\end{equation}
The consensus subset is $S=\{i:\hat{y}_i\neq\bot\}$, and ensemble coverage is defined as $|S|/N$, where membership in $S$ requires agreement on both pain fields. This rule prioritizes precision over recall by restricting downstream analyses to instances on which all three models agree.

For demographic attributes, we use majority voting over non-null predictions, retaining a value when at least two models agree and leaving the field missing when the predictions conflict without a majority. Thus, a value produced by only one model is retained only when no other model provides a conflicting prediction. Anatomical location and pain descriptors are represented as lists and aggregated item by item under the same two-of-three agreement rule. If only one model provides a list, its predictions are retained; otherwise, an item is retained only if at least two models include it. This strategy provides higher coverage for sparse attributes, such as ethnicity and occupation, for which requiring unanimous agreement across all three models would result in substantial data loss.

\paragraph{The abstain subset.}\label{app:abstain}
The strict agreement rule sets aside 7--11\% of WorldMedQA-Pain rows and up to 31.8\% of Brateca. Because WorldMedQA-Pain provides clinician-annotated labels for every row, including abstained rows, we can assess whether the abstain subset is systematically harder than the consensus subset. Table~\ref{tab:abstain} reports macro-F1 for both pain fields on the abstain subset for each individual model, the 2-of-3 majority vote, and GPT-5, with the 0.92--0.99 performance on the consensus subset serving as a reference. Difficulty is concentrated in pain presence: no system exceeds 0.75, and GPT-5 reaches only 0.63--0.73, whereas pain relevance is more recoverable (GPT-5: 0.80--0.98). These results indicate that the disagreements leading to abstention are associated primarily with uncertainty about pain presence rather than pain relevance. Thus, the consensus subset is enriched for cases in which pain presence can be determined more reliably, rather than representing a random sample of the corpus.

\begin{table*}[t]
\centering
\small
\setlength{\tabcolsep}{5pt}
\begin{tabular}{lcccccccc}
\toprule
& \multicolumn{2}{c}{\textbf{Brazil}}
& \multicolumn{2}{c}{\textbf{Japan}}
& \multicolumn{2}{c}{\textbf{Spain}}
& \multicolumn{2}{c}{\textbf{UK}} \\
\cmidrule(lr){2-3}\cmidrule(lr){4-5}\cmidrule(lr){6-7}\cmidrule(lr){8-9}
\textbf{Model}
& \textit{Rel.} & \textit{Pres.} & \textit{Rel.} & \textit{Pres.}
& \textit{Rel.} & \textit{Pres.} & \textit{Rel.} & \textit{Pres.} \\
\midrule
Gemma-4-E4B                  & 0.68 & 0.46 & 0.81 & 0.50 & 0.69 & 0.62 & 0.79 & 0.56 \\
Qwen3-8B                     & 0.56 & 0.50 & 0.60 & 0.61 & 0.76 & 0.61 & 0.62 & 0.53 \\
DeepSeek-R1-Distill-Qwen-32B & 0.46 & 0.56 & 0.49 & \textbf{0.75} & 0.53 & 0.47 & 0.62 & \textbf{0.75} \\
\midrule
2-of-3 majority vote         & 0.60 & 0.53 & 0.70 & 0.71 & 0.81 & \textbf{0.68} & 0.75 & 0.73 \\
GPT-5$^\dag$                 & \textbf{0.80} & \textbf{0.64} & \textbf{0.98} & 0.73 & \textbf{0.85} & \textbf{0.68} & \textbf{0.83} & 0.63 \\
\bottomrule
\end{tabular}
\caption{Pain macro-F1 on the non-consensus, or abstain, rows of
WorldMedQA-Pain, for each model, the 2-of-3 majority vote, and
GPT-5$^\dag$. All systems score below the consensus macro-F1 of 0.92
to 0.99 in Table~\ref{tab:bench}, most on pain-presence. Best per
column in \textbf{bold}; $^\dag$proprietary reference.}
\label{tab:abstain}
\end{table*}

\begin{table*}[t]
\centering
\small
\setlength{\tabcolsep}{8pt}
\begin{tabular}{llllll}
\toprule
\textbf{Field} & \textbf{Task} & \textbf{Definition} & \textbf{Input} & \textbf{Output} & \textbf{Metric} \\
\midrule
Pain relevance & IE & Pain-related concepts          & Text & True / False  & Macro-F1 \\
Pain presence  & IE & Pain assertion in the patient  & Text & True / False  & Macro-F1 \\
Age            & IE & Patient's age                  & Text & Number        & Accuracy \\
Gender         & IE & Patient's gender               & Text & Male / Female & Accuracy \\
Occupation     & IE & Patient's occupation           & Text & Text          & Accuracy \\
\bottomrule
\end{tabular}
\caption{The PAINT extraction pipeline, with each field's task, definition, input, output, and metric. IE: information extraction.}
\label{tab:tasks}
\end{table*}

\paragraph{Normalisation.}
We apply a deterministic post-processing pipeline to normalize demographic values across languages. Age is represented in years by the first numeric value in the extracted age field; ages under one year are recorded as 0. Gender is mapped to the binary labels \textit{male} and \textit{female} using a cross-lingual dictionary. Ethnicity and occupation are preserved as free text. Values that cannot be normalized are recorded as null.

\section{Clinician Validation}
\label{sec:validation}

The extracted annotations were validated by ten clinicians, two per dataset. For each BRIDGE-Pain dataset, each of its two annotators independently reviewed a random sample of 100 records, and for each WorldMedQA-Pain partition, 30 records.

For WorldMedQA-Pain, clinician validation covered the demographic attributes of age, gender, and occupation, while pain relevance and pain presence were evaluated against the clinician-annotated labels. For BRIDGE-Pain, clinicians validated pain relevance, pain presence, age, gender, and occupation; anatomical location, pain descriptors, and ethnicity were not validated. To assess inter-annotator reliability, the two annotators assigned to each partition independently annotated an overlapping subset of 50 BRIDGE-Pain records and 10 WorldMedQA-Pain records, without communication or access to each other's labels. Annotation dates for all annotators are reported in Appendix~\ref{app:provenance}.

Metrics are defined under Tasks and metrics below. A blank value confirmed as correctly absent counts as correct, and agreement between clinicians requires an exact match of normalised values, with two blanks counted as a match. Percentage agreement is used for demographic fields because high-cardinality or single-category samples can make $\kappa$ unstable or undefined; demographic $\kappa$ values and the matching rules are given in Appendix~\ref{app:clinval}.

As shown in Table~\ref{tab:clinval}, the ensemble achieves macro-F1 scores of 0.91--1.00 on BRIDGE-Pain, comparable to its 0.92--0.99 performance on WorldMedQA-Pain. Demographic extraction reaches 92--100\% accuracy across all five clinical datasets. Inter-annotator agreement is similarly high, with $\kappa=0.88$--1.00 for pain relevance, $\kappa=0.87$--1.00 for pain presence, and 90--100\% agreement for demographic attributes across the BRIDGE-Pain and WorldMedQA-Pain overlaps. Wherever defined, demographic $\kappa$ values range from 0.82 to 1.00, corresponding to almost-perfect agreement under the \citet{landis1977kappa} scale. Detailed per-field results, confusion matrices, and all observed disagreements are provided in Appendix~\ref{app:clinval}.

\begin{table}[t]
\centering
\small
\setlength{\tabcolsep}{4pt}
\begin{tabular}{lcccccc}
\toprule
& \multicolumn{2}{c}{\textbf{Pain}}
& \multicolumn{3}{c}{\textbf{Demographics}} \\
\cmidrule(lr){2-3}\cmidrule(lr){4-6}
\textbf{Dataset} & \textit{Rel.} & \textit{Pres.}
  & \textit{Age} & \textit{Gen.} & \textit{Occ.} \\
\midrule
\multicolumn{6}{l}{\textbf{(a) Ensemble vs.\ clinician} {\footnotesize (F1 / acc)}} \\
\cmidrule(lr){2-6}
MEDIQA    & 0.91 & 0.96 &  99 &  99 &  99 \\
MIMIC-III & 0.95 & 0.96 &  99 & 100 &  96 \\
Cantemist & 0.98 & 1.00 & 100 &  97 &  97 \\
IFMIR     & 1.00 & 1.00 &  92 &  95 &  97 \\
Brateca   & 1.00 & 0.97 &  97 & 100 & 100 \\
\midrule
\multicolumn{6}{l}{\textbf{(b) Clinician vs.\ clinician} {\footnotesize ($\kappa$ / \%)}} \\
\cmidrule(lr){2-6}
MEDIQA    & 0.88 & 1.00 & 100 & 100 &  98 \\
MIMIC-III & 0.88 & 0.92 & 100 & 100 &  98 \\
Cantemist & 1.00 & 1.00 & 100 & 100 & 100 \\
IFMIR     & 0.92 & 1.00 & 100 & 100 & 100 \\
Brateca   & 1.00 & 0.87 &  98 &  98 & 100 \\
\bottomrule
\end{tabular}
\caption{Clinician validation on BRIDGE-Pain, with 100 records per clinician
and a 50-record overlap per dataset. Panel (a): ensemble versus clinician annotations; macro-F1
is reported for pain relevance and pain presence, and accuracy for
demographic extraction. Panel (b): agreement between the two clinicians
on the overlapping subset; Cohen's $\kappa$ is reported for the binary
pain fields and percentage agreement for demographic fields.}
\label{tab:clinval}
\end{table}

\paragraph{Tasks and metrics.}\label{app:tasks}
PAINT is formulated as a multilingual information extraction benchmark. Table~\ref{tab:tasks} lists each clinician-validated field with its task, definition, input, output, and evaluation metric. We report macro-F1 for the binary pain tasks, the unweighted mean over the True and False classes of $F_1 = \frac{2PR}{P+R}$, and accuracy for demographic extraction, the fraction of records whose extracted value matches the clinician-confirmed value. Inter-annotator agreement is measured using Cohen's $\kappa = \frac{p_o - p_e}{1 - p_e}$ for pain relevance and pain presence, where $p_o$ is observed agreement and $p_e$ the agreement expected from the two annotators' marginals, and percentage agreement ($p_o$) for demographic attributes.

\begin{table*}[!t]
\centering
\small
\setlength{\tabcolsep}{4pt}
\begin{tabular}{llrrrrrrrr}
\toprule
\textbf{Split} & \textbf{Lang} & \textbf{$n$}
  & \textbf{PP-res.} & \textbf{Pain-pres.}
  & \textbf{Age} & \textbf{Gender} & \textbf{Ethnicity} & \textbf{Occupation} \\
\midrule
\multicolumn{9}{l}{\textit{WorldMedQA-Pain (exam corpus, $n{=}2{,}963$)}} \\
\cmidrule(lr){1-9}
Brazil REVALIDA   & pt &   904 &   804 (88.9\%) &   271 (33.7\%) & 83.1\% & 84.6\% &  1.9\% &  9.0\% \\
Japan NLE         & ja &   742 &   689 (92.9\%) &   112 (16.3\%) & 49.8\% & 49.9\% &  0.6\% &  5.2\% \\
Spain MIR         & es & 1{,}049 &   969 (92.4\%) &   136 (14.0\%) & 44.3\% & 42.6\% &  0.4\% &  2.0\% \\
UK PLAB           & en &   268 &   244 (91.0\%) &    90 (36.9\%) & 95.5\% & 98.4\% &  0.8\% &  2.1\% \\
\cmidrule(lr){1-9}
\multicolumn{9}{l}{\textit{BRIDGE-Pain (clinical text, $n{=}17{,}795$)}} \\
\cmidrule(lr){1-9}
MEDIQA chat-A     & en & 1{,}501 & 1{,}402 (93.4\%) &   286 (20.4\%) & 14.1\% & 40.2\% &  3.0\% &  7.1\% \\
MIMIC-III Mort.   & en & 5{,}000 & 3{,}993 (79.9\%) & 2{,}542 (63.7\%) & 87.0\% & 100.0\% & 89.6\% & 41.5\% \\
Cantemist-NER     & es & 1{,}294 & 1{,}091 (84.3\%) &   740 (67.8\%) & 95.6\% & 98.6\% &  1.9\% & 14.9\% \\
IFMIR-Incident    & ja & 5{,}000 & 4{,}744 (94.9\%) &   389 (8.2\%)  &  7.5\% &  4.2\% &  0.0\% &  3.2\% \\
Brateca-Mort.     & pt & 5{,}000 & 3{,}411 (68.2\%) & 2{,}392 (70.1\%) & 99.8\% & 99.9\% & 73.8\% &  7.9\% \\
\midrule
\textbf{Total}    &    & \textbf{20{,}758} & \textbf{17{,}347 (83.6\%)} & \textbf{6{,}958} & --- & --- & --- & --- \\
\bottomrule
\end{tabular}
\caption{Sample budget per split. \textbf{n}: extracted records.
\textbf{PP-res.}: rows where all three models agree on both pain
fields. \textbf{Pain-pres.}: pain-presence-positive rows within
PP-resolved. Demographic columns give ensemble coverage within the PP-resolved rows. The three
largest partitions are down-sampled to 5{,}000; Cantemist and MEDIQA
are in full.}
\label{tab:extraction-summary}
\end{table*}

\begin{table}[!t]
\centering
\footnotesize
\setlength{\tabcolsep}{2.5pt}
\begin{tabular}{lcccc}
\toprule
              & \textbf{Brazil} & \textbf{Japan} & \textbf{Spain} & \textbf{UK} \\
\midrule
Language      & pt          & ja          & es         & en   \\
Exam          & REVALIDA    & NLE         & MIR        & PLAB \\
Years         & 2011--23    & 2022--24    & 2018--23   & 2025 \\
$n$ items     & 904         & 742         & 1{,}049    & 268  \\
Pain-rel.~\% & 36.5        & 17.5        & 15.9       & 38.8 \\
Pain-pres.~\%& 30.0        & 15.1        & 13.0       & 33.6 \\
Location~\%  & 27.8        & 14.7        & 11.8       & 32.8 \\
\bottomrule
\end{tabular}
\caption{WorldMedQA-Pain per-partition statistics. Pain rates are
from the three-model ensemble; clinician-annotated labels only validate it, see
Appendix~\ref{app:annotation} and Table~\ref{tab:bench}. The Location
column is the share of items with a non-empty pain-location.}
\label{tab:wmqa-summary}
\end{table}

\section{Availability and Dataset Statistics}
\label{sec:repro}

\paragraph{Reproducibility.}
We release the extraction pipeline, prompts, schema, validation and agreement scripts, WorldMedQA-Pain labels, and clinician annotation sheets (source text removed) for the publicly licensed corpora; sheets for credentialed corpora are provided on request. The released consensus labels cover pain relevance and pain presence; anatomical location, pain descriptors, and demographic fields are provided as per-model raw outputs and can be aggregated with the released scripts. The clinician gold labels for WorldMedQA-Pain will be made available in a subsequent version. BRIDGE-Pain cannot be redistributed under the underlying corpus licenses, but users with access can regenerate the labels using the released pipeline. Every number in Tables~\ref{tab:clinval}, \ref{tab:clinval-wmqa}, \ref{tab:demo-kappa} and \ref{tab:confusion} reproduces from the sheets with the released scripts, as documented in Appendix~\ref{app:clinval}.

\paragraph{Dataset statistics.}\label{app:extraction-summary}\label{app:wmqa-summary}
Tables~\ref{tab:extraction-summary} and~\ref{tab:wmqa-summary} summarise the released labels per partition. The label distributions differ substantially between partitions and are not comparable prevalence estimates.

\section{Conclusion}
\label{sec:conclusion}

We introduced PAINT, to our knowledge the first multilingual resource for pain representation across medical licensing examinations and real-world clinical text in English, Spanish, Portuguese, and Japanese. It contributes clinician-validated pain and demographic annotations and the open-weight labeling pipeline.

PAINT is intended as a benchmark for multilingual clinical information extraction and as infrastructure for studying how pain documentation, and the demographic detail attached to it, varies across languages and care settings, including whether large language models reproduce known disparities in pain assessment. The pipeline extends to further languages and document types with minimal adaptation.

\section*{Limitations}

\paragraph{Cross-partition comparability.}
Corpus purpose, language, country, and healthcare system co-vary across the nine partitions, so the label distributions in Table~\ref{tab:extraction-summary} characterise each corpus rather than estimate prevalence. Differences between partitions should be read as corpus-purpose effects rather than as cultural claims about pain experience.

\paragraph{Scope.}
PAINT captures pain as documented by clinicians and examination writers, not as reported by patients; none of the source corpora carry numeric pain scores. The schema is designed to accommodate structured pain-intensity scores and patient-reported pain once corpora with such data become available.

\paragraph{Restricted data availability.}
The five BRIDGE-Pain partitions cannot be redistributed, so the clinical half of the resource must be regenerated by users with access to the source data, using the released pipeline.

\paragraph{Language-variety mismatch.}
Unlike the Spanish and Japanese partitions, the English partitions are not country-matched: WorldMedQA-Pain uses the UK PLAB examination, whereas BRIDGE-Pain contains US clinical corpora (MEDIQA and MIMIC-III), so exam-versus-clinical differences in English are confounded with British versus American clinical language. Also, both Portuguese partitions are Brazilian, so European Portuguese pain vocabulary is not represented.

\section*{Ethics Statement}

\paragraph{Data governance and human annotation.}
PAINT combines public medical licensing examinations with multilingual clinical corpora. WorldMedQA contains no patient-identifiable information, and the pain annotation study was granted IRB exemption by MIT. The annotation effort was conducted by ten clinicians (nine physicians and one pharmacist), all fluent in the source languages and listed as co-authors. Annotation procedures and timelines are documented in Appendix~\ref{app:provenance}.

\paragraph{Clinical data and redistribution.}
The five BRIDGE-Pain corpora are used in accordance with their original licensing and ethical frameworks, including the PhysioNet Credentialed Health Data License for MIMIC-III. PAINT does not redistribute source clinical text. Clinician annotation sheets are released with all source text removed for corpora whose licenses permit it (WorldMedQA, Cantemist, IFMIR); for the remaining corpora (MIMIC-III, MEDIQA, Brateca) they are provided on request to users with access to the underlying corpus.

\paragraph{Intended use.}
PAINT is designed to support multilingual clinical NLP research, information extraction, and fairness auditing. It is not intended for clinical diagnosis, treatment recommendation, or automated triage. Because demographic attributes may be sensitive, demographic results for restricted clinical datasets are reported as aggregate statistics; the only record-level values shown are the seventeen annotation disagreements in Appendix~\ref{app:clinval}, which carry no identifiers.

% Bibliography
\section*{Acknowledgements}
LAC is funded by the National Institutes of Health through DS-I Africa U54 TW012043-01 and Bridge2AI OT2OD032701, the National Science Foundation through ITEST \#2148451, a grant of the Boston-Korea Innovative Research Project (RS-2024-00403047), and a grant of the Korea Health Technology R\&D Project (RS-2024-00439677) through the Korea Health Industry Development Institute (KHIDI), funded by the Ministry of Health \& Welfare, Republic of Korea. YM was supported by the IVADO Postdoctoral Research Funding Program, funded through the Canada First Research Excellence Fund (CFREF).

\bibliography{medscope}

\clearpage
\appendix

\section{Compute and infrastructure}

 PAINT uses inference only, no model is trained or fine-tuned. The ensemble runs three open-weight models — Gemma-4-E4B (4B), Qwen3-8B (8B), and DeepSeek-R1-Distill-Qwen-32B (32B) — under 4-bit quantization, with GPT-5 (gpt-5-2025-08-07) included as a proprietary reference accessed through the OpenAI API. Open-weight inference was run on a single NVIDIA A100 (40 GB) or L4 (24 GB) GPU, with 4-bit NF4 quantization via bitsandbytes. GPT-5 was queried via API and used no local compute.

\section{Pain descriptor vocabulary by language}
\label{app:pain-vocab}

Table~\ref{tab:pain-expression} lists frequent tokens in the pain descriptors field (\texttt{pain\_type}) per language across both corpora. Japanese descriptors are compounds ending in {\jpfont 痛} that fuse location and quality in one word, whereas English, Spanish, and Portuguese use location-independent adjectives (\emph{sharp}, \emph{punzante}, \emph{aguda}) that translate almost directly among themselves.

\begin{table}[H]
\centering
\footnotesize
\setlength{\tabcolsep}{2pt}
\begin{tabular}{p{0.04\columnwidth} p{0.88\columnwidth}}
\toprule
\textbf{L.} & \textbf{Frequent pain descriptor tokens (count across WorldMedQA-Pain + BRIDGE-Pain)} \\
\midrule
en & pressure (153), sharp (129), discomfort (125), radiating (120),
     tenderness (107), intermittent (92), constant (62),
     burning (54), dull (43) \\[2pt]
pt & \emph{aguda} `sharp/acute' (227), \emph{desconforto}
     `discomfort' (106), \emph{dor} `pain' (72), \emph{intermitente}
     `intermittent' (65), \emph{cefaleia} `headache' (57),
     \emph{contínua} `continuous' (53), \emph{difusa} `diffuse'
     (43), \emph{irradia} `radiates' (41), \emph{opressiva /
     pressão} `oppressive / pressure' (41) \\[2pt]
es & \emph{irradiado/a} `radiating' (31), \emph{intermitente} `intermittent' (25), \emph{irradiado} `radiating' (18), \emph{c\'olico} `colicky' (16), \emph{continuo/a} `continuous' (14), \emph{continuo} `continuous' (14), \emph{neurop\'atico} `neuropathic' (13), \emph{sordo} `dull' (11), \emph{mec\'anico} `mechanical' (11), \emph{punzante} `stabbing' (10) \\
\addlinespace[2pt]
ja & {\jpfont 圧痛} \emph{tenderness}, lit.\ pressure-pain (42); {\jpfont 頭痛} \emph{headache}, lit.\ head-pain (21); {\jpfont 腹痛} \emph{abdominal pain} (18); {\jpfont 痛み} \emph{pain} (13); {\jpfont 咽頭痛} \emph{sore throat}, lit.\ throat-pain (10); {\jpfont 疼痛} \emph{ache/pain} (7); {\jpfont 関節痛} \emph{joint pain} (6); {\jpfont 血管痛} \emph{vessel pain} (6); {\jpfont 圧迫感} \emph{pressure-sensation} (5); {\jpfont 筋肉痛} \emph{muscle pain} (5) \\
\bottomrule
\end{tabular}
\caption{Frequent pain descriptor tokens per language, across both
corpora. Japanese descriptors fuse with the anatomical site or
pain-stem; the other languages use an independent quality channel.
Counts for BRIDGE-Pain are aggregate statistics only, as the
underlying text is not redistributed.}
\label{tab:pain-expression}
\end{table}

\section{Age composition per dataset}
\label{app:age-composition}

\begin{figure*}[t]
\centering
\includegraphics[width=\textwidth]{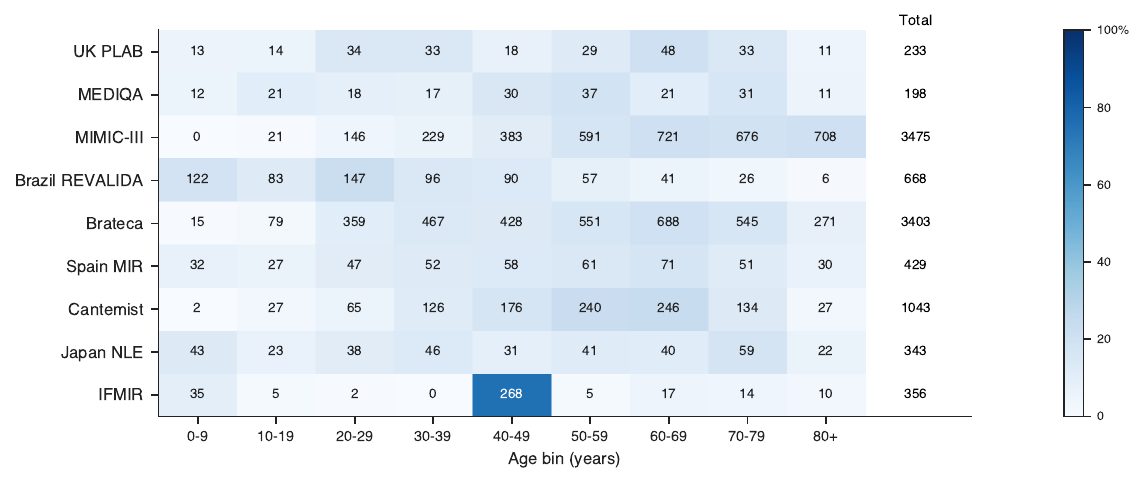}
\caption{Age composition per dataset. Each cell shows the bin's
sample size; colour encodes its share of the dataset, and rows sum to
100\%. \textbf{Total} is the age-labelled count within the pain-consensus subset.}
\label{fig:age-composition}
\end{figure*}

Figure~\ref{fig:age-composition} shows the age composition of each partition after normalisation, so that both the sample size of each bin and the age skew of each source are visible. The number in each cell is the bin's sample size, and its colour encodes the bin's share of the partition. The extreme bins are thin on several partitions, which users should keep in mind when stratifying by age.

\section{Prompts}
\label{app:prompts}

\begin{figure}[!t]
\centering
\includegraphics[width=\columnwidth]{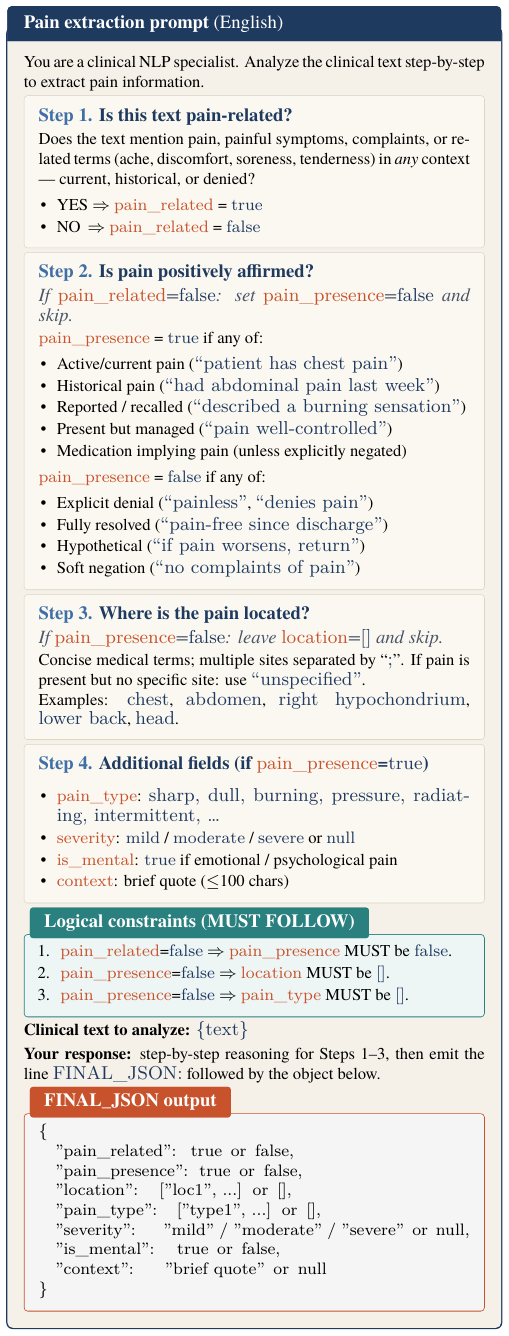}
\caption{Pain extraction prompt, English. The same four-step
template, hierarchical constraints, and \texttt{FINAL\_JSON} schema
are reused across all four languages; only the in-language lexicon
and worked examples differ, see
Figs.~\ref{fig:prompt:pain:jp}--\ref{fig:prompt:pain:pt}.}
\label{fig:prompt:pain}
\end{figure}

\begin{figure}[!t]
\centering
\includegraphics[width=\columnwidth]{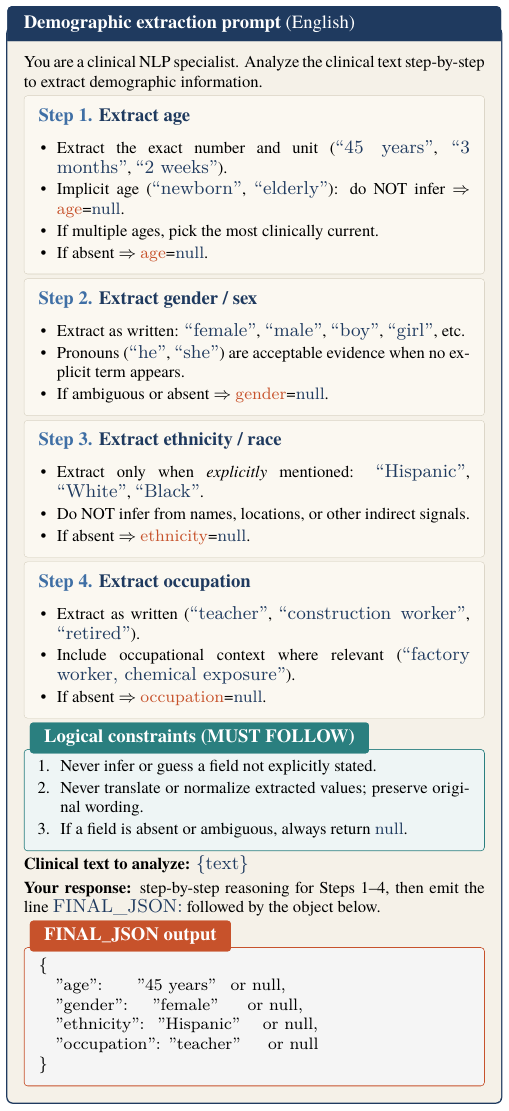}
\caption{Demographic extraction prompt, English. The same template
and never-infer policy are reused in the native-language variants
with locally-attested surface forms substituted in.}
\label{fig:prompt:demographic}
\end{figure}

\begin{figure}[!t]
\centering
\includegraphics[width=\columnwidth]{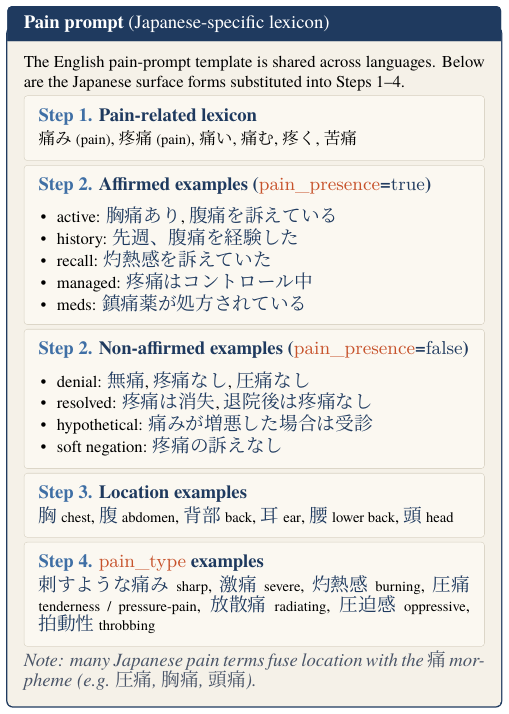}
\caption{Japanese pain-prompt substitutions: Step~1 lexicon and
Step~2--4 worked examples. The rest of the template,
Fig.~\ref{fig:prompt:pain}, is unchanged.}
\label{fig:prompt:pain:jp}
\end{figure}

\begin{figure}[!t]
\centering
\includegraphics[width=\columnwidth]{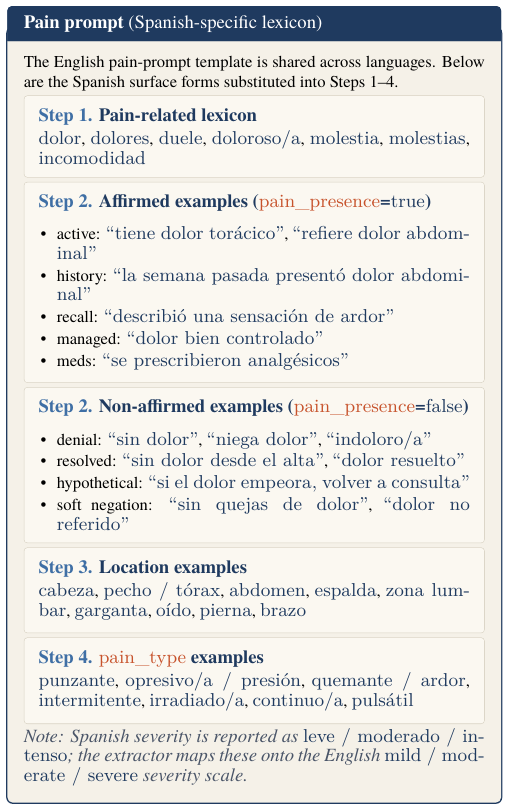}
\caption{Spanish pain-prompt substitutions.}
\label{fig:prompt:pain:es}
\end{figure}

\begin{figure}[!t]
\centering
\includegraphics[width=\columnwidth]{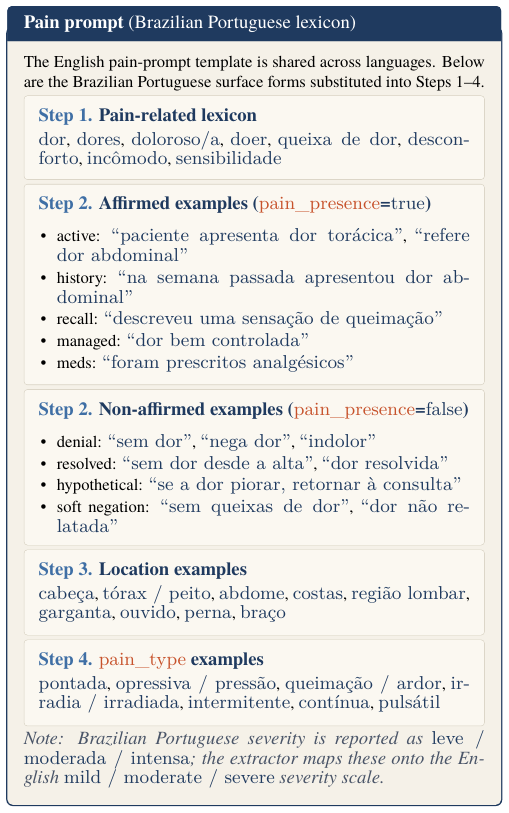}
\caption{Brazilian Portuguese pain-prompt substitutions.}
\label{fig:prompt:pain:pt}
\end{figure}

\begin{figure}[!t]
\centering
\includegraphics[width=\columnwidth]{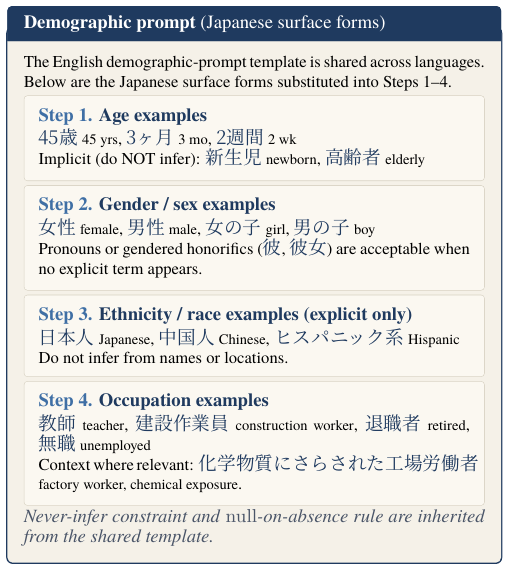}
\caption{Japanese demographic-prompt substitutions.}
\label{fig:prompt:demographic:jp}
\end{figure}

\begin{figure}[!t]
\centering
\includegraphics[width=\columnwidth]{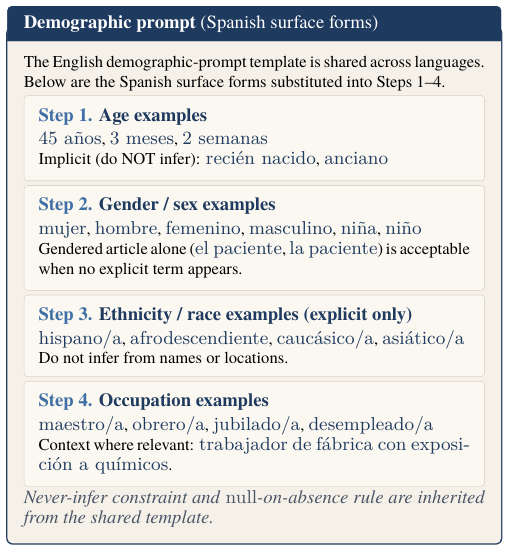}
\caption{Spanish demographic-prompt substitutions.}
\label{fig:prompt:demographic:es}
\end{figure}

\begin{figure}[!t]
\centering
\includegraphics[width=\columnwidth]{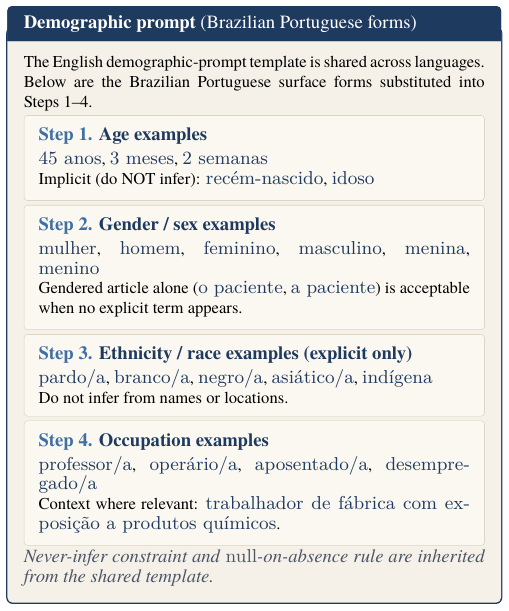}
\caption{Brazilian Portuguese demographic-prompt substitutions.}
\label{fig:prompt:demographic:pt}
\end{figure}

We list the prompt templates used in PAINT.
Figures~\ref{fig:prompt:pain} and \ref{fig:prompt:demographic} give
the English pain and demographic extraction prompts.
Figures~\ref{fig:prompt:pain:jp}, \ref{fig:prompt:pain:es}, and
\ref{fig:prompt:pain:pt} list the Japanese, Spanish, and Brazilian
Portuguese substitutions for the pain template, and
Figures~\ref{fig:prompt:demographic:jp},
\ref{fig:prompt:demographic:es}, and
\ref{fig:prompt:demographic:pt} do the same for the demographic
template.
The token \texttt{\{text\}} is the placeholder for the input
clinical text.

\paragraph{Prompt authorship.}
The English prompts were drafted by the authors against the
annotation schema and the protocol in Appendix~\ref{app:annotation}. The Japanese, Spanish, and Brazilian
Portuguese variants reuse the four-step English template verbatim;
only the lexicon entries, presence/non-presence worked examples, location labels, severity labels where applicable, and \texttt{pain\_type} descriptors are substituted with native surface forms. Each language's substitutions were authored by an
author or collaborator fluent in that language and cross-checked
against the per-language pain lexicon documented in
Table~\ref{tab:lexicons}.

\section{Clinician validation}
\label{app:clinval}

This appendix gives the per-field detail behind the clinician
validation summarised in Table~\ref{tab:clinval}; all numbers reproduce from the annotation sheets, released for the publicly licensed corpora and available on request for the credentialed ones, via \texttt{src/paint/clinician\_validation.py} and \texttt{src/paint/kappa\_transparency.py}.
Table~\ref{tab:clinval-ci} pools the BRIDGE-Pain validation across the
five datasets per field, on the records where the two clinicians agree;
disagreements are excluded and are reflected by $\kappa$ in
Table~\ref{tab:clinval}. It reports macro-F1 against clinician annotation for the binary pain
fields, and clinician-confirmed accuracy with a Wilson $95\%$ CI for
the extraction-style demographic fields. Pain macro-F1 is $0.98$ on
both fields and every demographic field clears $97\%$, so the released
silver labels track clinician judgement on the consensus subset at
high precision.

\paragraph{Matching rules.}
Accuracy in panel (a) of Table~\ref{tab:clinval} is the fraction of
records on which the clinician marked the ensemble value correct. The
judgement is the clinician's, not a string comparison: following the
instructions in Appendix~\ref{app:verification}, a blank ensemble value
is marked correct when the text states no such information, and an
occupation is marked correct when it names the occupation stated in
the text, so wording variants such as ``nurse'' and ``registered
nurse'' are not penalised. Percentage agreement in panel (b) compares
the two clinicians' final labels, each reconstructed as the ensemble
value when marked correct and as the supplied correction otherwise.
Labels are normalised before comparison: blank and null-equivalent
cells map to a single missing value, age to its first integer, and
gender to \emph{male} or \emph{female}; occupation is compared as an
exact string after lower-casing and whitespace trimming, so ``nurse''
and ``registered nurse'' count as a disagreement here. Two missing
values count as agreement. Records on which either clinician left the
field unjudged are excluded from the agreement denominator.

Table~\ref{tab:demo-kappa} gives, for every demographic field,
percent agreement alongside Cohen's $\kappa$ and the chance-agreement
term $p_e$ that $\kappa$ corrects for, so that readers can see where
$\kappa$ is informative and where it is undefined.
Table~\ref{tab:disagreements} lists every disagreement observed across
all double-annotated overlaps, and Table~\ref{tab:confusion} gives the
Cantemist confusion matrices in full. Cantemist consists of structured oncology case reports in which pain and demographics are stated explicitly, which may account for the near-perfect agreement. Each Cantemist annotator labelled 100 records with a 50-record overlap (150 unique records in total), and the two agree on every overlapping record ($\kappa=1.00$ on all five fields); the same clinician also annotated the WorldMedQA Spain-MIR partition, where agreement is 100\%.
Table~\ref{tab:clinval-wmqa} reports the WorldMedQA-Pain
\emph{demographic} validation, which we keep out of the main results:
the exam pain fields already carry clinician-annotated labels,
reported in Table~\ref{tab:bench}, and exam demographics, stated explicitly in
the vignette, are near-ceiling, so this is a confirmatory check, not a
finding.

\begin{table}[h]
\centering
\small
\setlength{\tabcolsep}{4pt}
\begin{tabular}{llcc}
\toprule
\textbf{Field} & \textbf{Metric} & \textbf{Value} & \textbf{95\% CI} \\
\midrule
Pain-relevance & macro-F1 & 0.98   & --- \\
Pain-presence  & macro-F1 & 0.98   & --- \\
Age            & accuracy & 97.3\% & [95.9, 98.3] \\
Gender         & accuracy & 98.1\% & [96.9, 98.9] \\
Occupation     & accuracy & 97.9\% & [96.6, 98.7] \\
\bottomrule
\end{tabular}
\caption{BRIDGE-Pain validation pooled across the five datasets, on
records where the two clinicians agree, about 750 each. Pain fields
give the ensemble macro-F1 against clinician annotation; demographic
fields give clinician-confirmed accuracy with a Wilson 95\% confidence
interval. Per-dataset values are in Table~\ref{tab:clinval}.}
\label{tab:clinval-ci}
\end{table}

\begin{table}[h]
\centering
\small
\setlength{\tabcolsep}{4pt}
\begin{tabular}{lccc}
\toprule
\textbf{Split} & \textit{Age} & \textit{Gen.} & \textit{Occ.} \\
\midrule
\multicolumn{4}{l}{\textit{Clinician-confirmed accuracy (\%)}} \\
\cmidrule(lr){2-4}
UK PLAB         & 100 & 100 & 100 \\
Spain MIR       & 100 &  96 & 100 \\
Japan NLE       & 100 & 100 & 100 \\
Brazil REVALIDA &  92 &  98 &  96 \\
\midrule
\multicolumn{4}{l}{\textit{Inter-annotator agreement (\%)}} \\
\cmidrule(lr){2-4}
UK PLAB         &  90 & 100 & 100 \\
Spain MIR       & 100 & 100 & 100 \\
Japan NLE       & 100 & 100 & 100 \\
Brazil REVALIDA & 100 &  90 & 100 \\
\bottomrule
\end{tabular}
\caption{WorldMedQA-Pain demographic validation, 30 records per
partition with an overlap of 10 for agreement. Pain fields carry
clinician-annotated labels, Table~\ref{tab:bench}, and are not re-annotated here.
Accuracy is over the 50 unique records per partition. Cohen's $\kappa$ on the 10-record overlaps, where defined, is 0.82--1.00, lowest for Brazil REVALIDA gender (0.82) and UK PLAB age (0.89).}
\label{tab:clinval-wmqa}
\end{table}

\begin{table}[h]
\centering
\small
\setlength{\tabcolsep}{3pt}
\begin{tabular}{llrrrr}
\toprule
\textbf{Dataset} & \textbf{Field} & \textbf{\#cat} & \textbf{\%} & $\bm{p_e}$ & $\bm{\kappa}$ \\
\midrule
MEDIQA    & age        &  6 & 100 & 0.81 & 1.00 \\
          & gender     &  3 & 100 & 0.52 & 1.00 \\
          & occupation &  6 &  98 & 0.83 & 0.88 \\
MIMIC-III & age        & 30 & 100 & 0.07 & 1.00 \\
          & gender     &  2 & 100 & 0.50 & 1.00 \\
          & occupation & 21 &  98 & 0.30 & 0.97 \\
Cantemist & age        & 35 & 100 & 0.04 & 1.00 \\
          & gender     &  2 & 100 & 0.51 & 1.00 \\
          & occupation &  6 & 100 & 0.81 & 1.00 \\
IFMIR     & age        &  2 & 100 & 0.96 & 1.00 \\
          & gender     &  1 & 100 & 1.00 & --- \\
          & occupation &  1 & 100 & 1.00 & --- \\
Brateca   & age        & 37 &  98 & 0.03 & 0.98 \\
          & gender     &  3 &  98 & 0.53 & 0.96 \\
          & occupation &  4 & 100 & 0.88 & 1.00 \\
\bottomrule
\end{tabular}
\caption{Demographic inter-annotator agreement on the 50-record
BRIDGE-Pain overlaps: number of distinct categories, percent
agreement, expected chance agreement $p_e$, and Cohen's $\kappa$.
$\kappa$ is undefined (---) where the overlap contains a single
category, so $p_e{=}1$; it is near-uninformative where $p_e$ is close
to 1 (IFMIR age). For IFMIR gender and occupation, both clinicians record every record in the 50-record overlap as null and $\kappa$ is undefined on a single-category field. Where $p_e$ is small, as for age on MIMIC-III,
Cantemist, and Brateca with 30--37 distinct values, perfect agreement
is the condition under which $\kappa$ is \emph{least} inflated.}
\label{tab:demo-kappa}
\end{table}

\begin{table}[h]
\centering
\small
\setlength{\tabcolsep}{3pt}
\begin{tabular}{p{1.9cm}p{1.9cm}p{1.3cm}p{1.5cm}}
\toprule
\textbf{Dataset} & \textbf{Field} & \textbf{Clin.\ A} & \textbf{Clin.\ B} \\
\midrule
MEDIQA  & pain-relevance & false & true \\
MEDIQA  & pain-relevance & false & true \\
MEDIQA  & occupation     & null  & registered nurse \\
MIMIC-III & pain-relevance & false & true \\
MIMIC-III & pain-relevance & false & true \\
MIMIC-III & pain-relevance & false & true \\
MIMIC-III & pain-presence  & false & true \\
MIMIC-III & pain-presence  & false & true \\
MIMIC-III & occupation   & nurse & former nurse \\
Brateca & pain-presence  & true  & false \\
Brateca & pain-presence  & true  & false \\
Brateca & pain-presence  & true  & false \\
Brateca & age            & 38    & null \\
Brateca & gender         & male  & null \\
IFMIR & pain-relevance & false & true \\
UK PLAB & age & 0 & null \\
Brazil \mbox{REVALIDA} & gender & null  & male \\
\bottomrule
\end{tabular}
\caption{Every disagreement across all double-annotated overlaps
($n{=}17$ of $1{,}370$ judgements). MEDIQA pain-relevance disagreements arise on doctor--patient dialogue turns where intent must be judged; the MIMIC-III pain disagreements concern whether pain elicited during a neurological examination counts as a pain mention; the IFMIR case turns on whether {\jpfont 気分不快} (malaise) is a pain term; Brateca pain-presence disagreements concern assertion strength. The Brateca age/gender pair is one 9{,}400-character
record whose header states \texttt{Age: 38 \ldots\ Sex:} explicitly;
clinician B marked both fields as not stated, a reading slip on a
long record rather than schema ambiguity. In the UK PLAB age case, clinician B confirmed the empty extraction and clinician A entered 0. No judgement was altered
after the fact.}
\label{tab:disagreements}
\end{table}

\begin{table}[h]
\centering
\small
\setlength{\tabcolsep}{4pt}
\begin{tabular}{lrrr}
\toprule
\textbf{Field} & \textbf{Agree} & \textbf{Disagree} & \textbf{Marginals (A)} \\
\midrule
pain-relevance & 50 & 0 & 38 true / 12 false \\
pain-presence  & 50 & 0 & 35 true / 15 false \\
gender         & 50 & 0 & 29 female / 21 male \\
age            & 50 & 0 & 35 distinct values \\
occupation     & 50 & 0 & 5 stated / 45 null \\
\bottomrule
\end{tabular}
\caption{Cantemist confusion matrices on the 50-record overlap, all
diagonal. Marginals are close to balanced for the binary fields
($p_e{=}0.51$ for gender, essentially the minimum for a binary
field) and negligible for age, so the agreement is not an artefact
of a degenerate label distribution. Cantemist consists of Spanish
oncology case reports written to a rigid template in which age and
sex are stated verbatim in the opening clause (``Mujer de 45 años
sin alergias\ldots''), leaving nothing to infer. Agreement between
annotators does not mean agreement with the ensemble: both clinicians
independently rejected three ensemble values (record 1147
pain-relevance false$\to$true; 1217 gender female$\to$male; 949
occupation ``exfumadora''$\to$none), which is why panel (a) of
Table~\ref{tab:clinval} is below 1.00 where panel (b) is not.}
\label{tab:confusion}
\end{table}

\section{Verification protocol and annotator instructions}
\label{app:verification}

WorldMedQA-Pain pain labels are produced by open annotation under the
three-step procedure in Appendix~\ref{app:annotation}. All other
clinician labels, the five BRIDGE-Pain fields and the WorldMedQA-Pain
demographic fields, are produced by \emph{verification}: the annotator
sees the ensemble's value for each field and judges whether it is
correct, supplying the correct value when it is not. The instructions below were sent to every annotator with their sheet and are abridged here to the parts that define the task; the WorldMedQA-Pain version listed only the three demographic fields.

\begin{quote}\small
We have used language models to extract pain and demographic
information from clinical text, and we need clinician judgment to
verify how accurate those extractions are. Your task is to mark whether
each extracted value is correct. Each row of the spreadsheet shows a
value our system extracted (the \texttt{\ldots\_\_consensus} column).
For each field, please fill in two columns:
\texttt{\ldots\_\_correct(TRUE/FALSE)}, TRUE if the value is correct,
FALSE if not; \texttt{\ldots\_\_true\_value\_if\_false}, only when you
marked FALSE: write the correct value. Please read the clinical text in
the \texttt{input\_text} column to decide, and do not edit the
\texttt{\_\_consensus} or \texttt{input\_text} columns.

Fields you'll see: \texttt{pain\_related}, the text mentions pain in
any way (including pain that is denied/ruled out);
\texttt{pain\_presence}, pain is affirmed as actually present (now or
in the past); \texttt{age}, \texttt{gender}, \texttt{occupation},
patient demographics as stated in the text.

One important point: a blank consensus value means our system found no
such information in the text. If the text contains no age/gender/etc.,
mark TRUE; if the information is there and we missed it, mark FALSE
and supply the correct value.
\end{quote}

Each sheet holds one row per record. Every row begins with identifier
columns and the source text, then carries three columns per field,
listed in Table~\ref{tab:sheetcols}. BRIDGE-Pain sheets carry
\texttt{row\_index, orig\_id, dataset, language, input\_text} and the
three columns for each of five fields, twenty columns in total;
WorldMedQA-Pain sheets carry \texttt{row\_index, question\_id, dataset,
language, input\_text} and the three columns for age, gender and
occupation only, fourteen in total.

\begin{table}[h]
\centering
\small
\setlength{\tabcolsep}{4pt}
\begin{tabular}{p{2.3cm}lp{2.9cm}}
\toprule
\textbf{Column} & \textbf{Filled by} & \textbf{Content} \\
\midrule
\texttt{<field>\_\_}\newline\texttt{consensus} & ensemble & silver label from the three-model consensus, pre-filled \\
\texttt{<field>\_\_}\newline\texttt{correct(TRUE/}\newline\texttt{FALSE)} & annotator & whether the ensemble value is correct \\
\texttt{<field>\_\_}\newline\texttt{true\_value\_}\newline\texttt{if\_false} & annotator & the correct value, supplied when FALSE \\
\bottomrule
\end{tabular}
\caption{Per-field columns of every annotation sheet.}
\label{tab:sheetcols}
\end{table}

A FALSE judgement with the correction left blank encodes ``the text
does not state this value''; it is a judgement, not a skipped cell. A
blank \texttt{\_\_correct} cell would be a skipped cell, treated as
missing data; none occur in the released sheets.

\section{Annotation provenance}
\label{app:provenance}

Table~\ref{tab:provenance} gives the recruitment and return date of
every clinician for every partition, taken from the assignment tracker
maintained throughout the project. Clinicians are coded P1--P10; the
same code denotes the same person across rows. P5, a pharmacist (all others are physicians), annotated the two Spanish
partitions in July; all other partitions were annotated in May. All annotations were carried out individually. Every annotator worked from a separate sheet with no
communication and no visibility into any other annotator's labels or
into whether their own judgements matched the pipeline's.

The four English annotators are physicians who trained in English-medium medical programmes outside the United Kingdom and the United States and for whom English is the working clinical language; at the time of annotation, three were in clinical training programmes in the United States and one in Canada. Annotators for Spanish, Portuguese, and Japanese are native speakers of the corpus language.

Annotators verified a displayed ensemble label against the source text
rather than labelling from scratch. This protocol has a consequence for
the agreement ceiling: where the displayed label is correct and the
text is explicit, two competent readers should agree, and high
agreement is the expected outcome rather than a surprising one.
Table~\ref{tab:disagreements} shows that the same protocol does
register disagreement where the text is genuinely ambiguous.

\begin{table}[h]
\centering
\small
\setlength{\tabcolsep}{3pt}
\begin{tabular}{llcll}
\toprule
\textbf{Partition} & \textbf{Lang} & \textbf{Clin.} & \textbf{Sent} & \textbf{Returned} \\
\midrule
MEDIQA          & en & P1  & 21 May & 21 May \\
MEDIQA          & en & P2  & 21 May & 22 May \\
MIMIC-III       & en & P3  & 21 May & 22 May \\
MIMIC-III       & en & P4  & 21 May & 22 May \\
Cantemist       & es & P5  & 9 Jul  & 10 Jul \\
Cantemist       & es & P6  & 21 May & 22 May \\
IFMIR           & ja & P7  & 21 May & 22 May \\
IFMIR           & ja & P8  & 21 May & 22 May \\
Brateca         & pt & P9  & 21 May & 25 May \\
Brateca         & pt & P10 & 21 May & 24 May \\
\midrule
UK PLAB         & en & P1  & 21 May & 21 May \\
UK PLAB         & en & P2  & 21 May & 22 May \\
Spain MIR       & es & P5  & 9 Jul  & 11 Jul \\
Spain MIR       & es & P6  & 21 May & 22 May \\
Japan NLE       & ja & P7  & 21 May & 21 May \\
Japan NLE       & ja & P8  & 21 May & 22 May \\
Brazil REVALIDA & pt & P9  & 21 May & 24 May \\
Brazil REVALIDA & pt & P10 & 21 May & 24 May \\
\bottomrule
\end{tabular}
\caption{Date each annotation sheet was sent to and returned by each
clinician, 2026. Upper block: BRIDGE-Pain; lower block:
WorldMedQA-Pain.}
\label{tab:provenance}
\end{table}

\section{Released spot-check protocol}
\label{app:spotcheck}

The two binary pain fields (pain-relevance and pain-presence) carry clinician validation across every
WorldMedQA-Pain row; the macro-F1
numbers in Table~\ref{tab:bench} summarise that agreement. The remaining fields (anatomical location, pain descriptors, severity, the extracted context span, ethnicity, and, beyond the samples in Appendix~\ref{app:clinval}, age, gender, and occupation) come from the ensemble alone.
The clinician validation in Appendix~\ref{app:clinval} follows the same verify-the-displayed-label design; the protocol below extends it to the ensemble-only fields, and we specify it here so that downstream users can audit those fields on their own samples without re-implementing the sampling, rating, and
reporting scheme.

\paragraph{Sampling frame.}
For each of the four languages (en, pt, es, ja), the protocol
draws a stratified random sample of 30--50 records per field whose
ensemble-resolved coverage is above 5\%:
\begin{itemize}\itemsep0pt
  \item pain-type, context (conditional on
        pain-presence true);
  \item age, gender, ethnicity, occupation (conditional on the
        field being ensemble-resolved).
\end{itemize}
Within each (language, field) cell the sample is balanced across
both corpora where both carry ensemble-resolved subsets.

\paragraph{Annotator task.}
A clinician annotator fluent in the source language reviews each
sampled record and rates the ensemble's output on a three-point
scale: \textsf{matches} (the ensemble value agrees with the source
text), \textsf{diverges} (the source text supports a different
value), or \textsf{ambiguous} (the source text leaves both readings
open). A free-text note accompanies each \textsf{diverges} or
\textsf{ambiguous} case.

\paragraph{Reported metrics.}
For each (language, field) cell the protocol reports
\textsf{matches}-rate with a Wilson 95\% confidence interval, the
ambiguous rate, and the top three failure modes from the free-text
notes.

\paragraph{Scope.}
The protocol measures ensemble accuracy conditional on the
ensemble resolving the row. Recall on the abstain subset (records
where the three models disagree) sits outside the protocol, and
findings in the paper are reported as conditional on
the ensemble-resolved population for the same reason.

\section{WorldMedQA-Pain annotation protocol}
\label{app:annotation}

This appendix documents the protocol used to produce the clinician
gold pain labels for WorldMedQA-Pain.
The protocol was developed once and instantiated per language; the per-language lexicons are listed in Table~\ref{tab:lexicons}.

\paragraph{Decision procedure.} Each exam item is annotated on its
native-language question text in two ordered steps, mirroring the
hierarchical pain schema:
\begin{enumerate}\itemsep0pt
  \item \textbf{pain-relevance.} Mark \textsf{Yes} if any
        pain-related concept appears anywhere in the item (a pain term,
        a synonym, or a morphological derivative such as
        ``painful'', ``painless'', or ``painkiller''),
        including explicit denial; otherwise \textsf{No}.
  \item \textbf{pain-presence.} For pain-relevance${=}%
        \textsf{Yes}$ items, mark \textsf{Yes} if pain is asserted in
        the patient at least once in the description; mark
        \textsf{No} if pain appears only as a denial or only as an
        attribute of an intervention (``painless'',
        ``painkiller'').
\end{enumerate}
Items that did not resolve cleanly under this procedure were highlighted
and adjudicated by the team rather than decided unilaterally.

\paragraph{Per-language pain lexicons.} Each language uses a lexicon
derived from an authoritative reference for that language and reviewed by
the per-country annotators. Representative entries are listed in
Table~\ref{tab:lexicons}. The lexicons are intentionally broad: they
cover base terms, morphological derivatives, the Greek
\mbox{\emph{-algia}} suffix family, and high-frequency medical compounds
(e.g.\ \emph{cefalea}/\emph{cefaleia} for headache). 

\begin{table}[t]
\centering
\footnotesize
\setlength{\tabcolsep}{3pt}
\renewcommand{\arraystretch}{1.15}
\begin{tabular}{@{}p{1.45cm}p{5.2cm}@{}}
\toprule
\textbf{Language} & \textbf{Representative lexicon entries} \\
\midrule
English \newline (Oxford) &
pain, ache, discomfort, soreness, hurt, agony, throb, sting, twinge, pang,
spasm, cramp; derivatives: painful, painless, painkiller \\
Spanish \newline (RAE) &
dolor (+ doloroso, adolorido, indoloro, \ldots); molestia, mal, aflicci\'on,
da\~no, tormento; mental sense: sufrimiento, angustia, congoja;
\mbox{-algia} family: neuralgia, mialgia, lumbalgia, \ldots; cefalea \\
Portuguese \newline (BR) &
dor (+ doloroso, dolorido, indolor, \ldots); machucado, desconforto, les\~ao,
c\'olica, c\~aimbra; mental sense: ang\'ustia, sofrimento; \mbox{-algia}
family: mialgia, neuralgia, lombalgia, \ldots; cefaleia \\
Japanese &
general: {\jpfont 激痛} (severe), {\jpfont 痛み}/{\jpfont 疼痛} (pain),
{\jpfont 鈍痛} (dull); localised: site + the bound morpheme {\jpfont 痛}
(e.g.\ {\jpfont 胸痛} chest, {\jpfont 腹痛} abdomen, {\jpfont 頭痛} head);
mental: {\jpfont 社会的苦痛}, {\jpfont 苦痛}, {\jpfont 気分不快};
physical-exam / denial: {\jpfont 圧痛}, {\jpfont 叩打痛},
{\jpfont 無痛性} \\
\bottomrule
\end{tabular}
\caption{Representative per-language pain-lexicon entries from the
annotation protocol, in source orthography. The full lexicons ship
with the code.}
\label{tab:lexicons}
\end{table}

\end{document}